\documentclass[letterpaper, 10 pt, journal, twoside]{IEEEtran}
\usepackage{algpseudocode}
\usepackage{algorithm}
\algnewcommand\algorithmicforeach{\textbf{for each:}}
\algnewcommand\ForEach{\item[ \algorithmicforeach]}

\newcommand{\baselinecolor}{orange }
\newcommand{\tactilechannelcolor}{blue }

\ifCLASSINFOpdf
  \usepackage[pdftex]{graphicx}
  \graphicspath{{figures/}}
\else
\fi
\usepackage{amsmath}
\usepackage{subcaption}
\begin{document}
%
\title{VisuoTactile 6D Pose Estimation of an In-Hand Object using Vision and Tactile Sensor Data}
%
%
%

\author{Snehal Dikhale$^{1}$, 
Karankumar Patel$^{1}$, 
Daksh Dhingra$^{3}$, 
Itoshi Naramura$^{2}$, 
Akinobu Hayashi$^{2}$,
Soshi Iba$^{1}$ 
and Nawid Jamali$^{1}$%
\thanks{Manuscript received: September, 9, 2021; Revised November, 25, 2021; Accepted December, 21, 2021.}
\thanks{This paper was recommended for publication by Markus Vincze upon evaluation of the Associate Editor and Reviewers' comments.} 
\thanks{$^{1}$Honda Research Institute USA, Inc.
       {\tt\footnotesize \{sdikhale, karankumar\_patel, siba, njamali\}@honda-ri.com}}
\thanks{$^{2}$Honda R\&D Co., Ltd. Innovative Research Excellence, Saitama, Japan
       {\tt\footnotesize \{itoshi\_naramura,akinobu\_hayashi\}@jp.honda}}
\thanks{$^{3}$Department of Mechanical Engineering, University of Washington, Seattle, USA. This work was done while he was an intern at Honda Research Institute USA, Inc.
       {\tt\footnotesize dd292@uw.edu}}%
\thanks{Digital Object Identifier (DOI): see top of this page.}
}
%
%

\markboth{IEEE Robotics and Automation Letters. Preprint Version. Accepted January, 2022}
{Dikhale \MakeLowercase{\textit{et al.}}: VisuoTactile 6D Pose Estimation of an In-Hand Object} 

%



\maketitle

\begin{abstract}
Knowledge of the 6D pose of an object can benefit in-hand object manipulation. Existing 6D pose estimation methods use vision data. In-hand 6D object pose estimation is challenging because of heavy occlusion produced by the robot's grippers, which can have an adverse effect on methods that rely on vision data only. Many robots are equipped with tactile sensors at their fingertips that could be used to complement vision data. In this paper, we present a method that uses both tactile and vision data to estimate the pose of an object grasped in a robot's hand. The main challenges of this research include 1) lack of standard representation for tactile sensor data, 2) fusion of sensor data from heterogeneous sources---vision and tactile, and 3) a need for large training datasets. To address these challenges, first, we propose use of point clouds to represent object surfaces that are in contact with the tactile sensor. Second, we  present a network architecture based on pixel-wise dense fusion to fuse vision and tactile data to estimate the 6D pose of an object. Third, we extend NVIDIA's Deep Learning Dataset Synthesizer to produce synthetic photo-realistic vision data and the corresponding tactile point clouds for 11 objects from the YCB Object and Model Set in Unreal Engine 4. We  present results of simulated experiments suggesting that using tactile data in addition to vision data improves the 6D pose estimate of an in-hand object. We also present qualitative results of experiments in which we deploy our network on real physical robots showing successful transfer of a network trained on synthetic data to a real system.
\end{abstract}


%
\IEEEpeerreviewmaketitle

\section{Introduction}
\label{sec:introduction}
%
%
%
%
\IEEEPARstart{A}{ccurate} estimate of 6D pose---3D rotation and 3D translation--- of an object can be used in tasks like manipulation and grasping, planning, and in virtual reality applications. Recent research in pose estimation applies deep learning to color and depth images to estimate the 6D pose of an object~\cite{Wang2019DenseFusion:Fusion,Tremblay2018DeepObjects,Xiang2018PoseCNN:Scenes}. Color images capture texture well ~\cite{Xiang2018PoseCNN:Scenes}, and depth data carries geometric information, enhancing the prediction accuracy under different lighting conditions~\cite{Wang2019DenseFusion:Fusion}.

Majority of research on 6D pose estimation is focused on  scenes partially occluded by other objects~\cite{Xiang2018PoseCNN:Scenes,Tremblay2018DeepObjects,Wang2019DenseFusion:Fusion}. However, 6D pose of an in-hand object has received little attention~\cite{bimbo2016hand}. Challenges of in-hand object pose estimation include heavy occlusion caused by the robot's gripper and other objects in the environment. In such situations, a robot can take advantage of its tactile sensors to improve its estimate. Furthermore, tactile data captures surface geometry that is otherwise occluded by the object itself~\cite{Watkins-Valls2019}. In this paper, we present a method that fuses tactile data with vision data to estimate the 6D pose of an object grasped in a robot's hand. We hypothesize that using both tactile and vision data will be an improvement over using vision data alone.

\begin{figure}
    \centering
    \includegraphics[clip,width=0.48\textwidth]{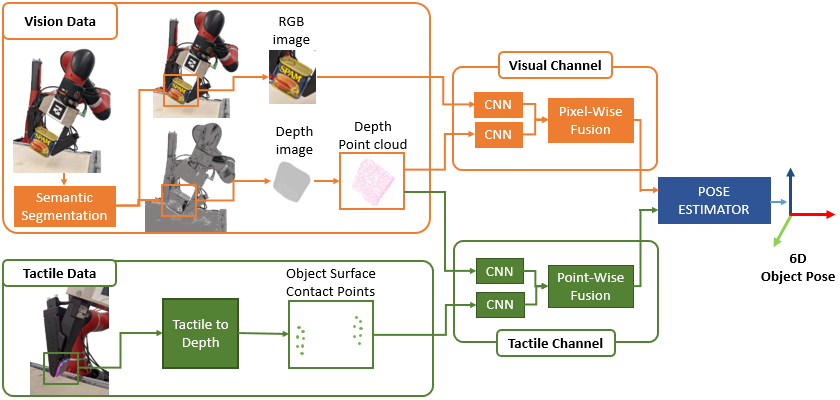}
    \caption{An illustration of the proposed method. The input to our model is color and depth images from a camera, and tactile contact points in the form of a point cloud. The output is the estimated 6D pose of the object of interest.\vspace{-0.4cm}}
    \label{fig:approach-overview}
\end{figure}

Figure~\ref{fig:approach-overview} is an illustration of the proposed method. The input to our model is color and depth image from a camera, and tactile contact points in the form of a point cloud. We use pixel-wise dense fusion proposed by Wang et al.~\cite{Wang2019DenseFusion:Fusion} to combine the sensor data. To allow meaningful fusion of heterogeneous sensor data such as vision and tactile, the network is divided into two channels: the visual channel and the tactile channel. In the visual channel, features from the color image and the point cloud from the depth image are fused at pixel level. In the tactile-channel, point cloud features from the depth image, and the object-surface point cloud features from the tactile sensors are fused at point level. The intuition behind this fusion is that the color and the depth images are correlated. Similarly, the object-surface point cloud from the tactile sensors and the point cloud from the depth images are correlated. Thus, the point cloud from the depth image connects the color image and the tactile senor data, accounting for the parts occluded by the robot's grippers.

Collecting training data for robotic manipulation is time-consuming and prohibitively expensive due to robot wear and tear. One approach is to use synthetic data to train a model. Synthetic data from vision sensors has been successfully applied to the problem of grasping~\cite{Mahler2017Dex-NetMetrics}, and 6D pose estimation~\cite{Tremblay2018DeepObjects}. 
However, synthetic data generation for manipulation that includes tactile data is less explored. Unlike vision sensors, tactile sensors are less developed\cite{Yamaguchi2019RecentVision}. There is no standard representation format for tactile data. The representation depends on the underlying transduction method, for example, optical-based tactile sensors output RBG images~\cite{Johnon2009GelSight}, while others may have randomly distributed sensors~\cite{Wettels2014BioTacs}. Thus, in this paper, we propose a representation format suitable for generating  synthetic tactile sensor data for 6D pose estimation. We represent tactile data as object-surface point cloud at the finger-object contact location. We extended NVIDIA's Deep Learning Dataset Synthesizer (NDDS) ~\cite{to2018ndds}, to produce synthetic photo-realistic vision data, the corresponding object-surface point clouds at the finger-object contact locations, and the ground truth 6D poses for 11~objects from the YCB Object and Model Set~\cite{Calli2015TheResearch}. Our dataset contains 20000 examples for each object.

The contributions of this paper include: 1) a tactile sensor invariant representation to capture finger-object contact surface, 2) a network architecture to fuse vision and tactile data to estimate the 6D pose of an in-hand object under heavy occlusion, and 3)
a method to generate synthetic visuotactile datasets suitable for in-hand object 6D pose estimation.

We present results of experiments in which we show our model, trained on the proposed synthetic visuotactile dataset, outperforms a baseline model trained using only the color and the depth images from the same dataset. We also present qualitative results of experiments in which our network and the baseline network trained on the synthetic dataset are deployed on real physical robots and show that the network generalizes to different physical setups.

\section{Background}
Existing $6$D pose estimation literature can be divided into three main categories: geometry matching~\cite{bimbo2016hand}, probabilistic methods~\cite{Aggarwal2014ObjectSensors,Fox2012TactileRobot,Vazquez2014In-handSensors} and machine learning methods~\cite{Varley2017ShapeGrasping,Watkins-Valls2019}. Geometric methods use tactile-geometric-shape coherence descriptors to estimate the pose of an object~\cite{bimbo2016hand}. These techniques suffer from slow run-time performance. Probabilistic methods use particle filters~\cite{Aggarwal2014ObjectSensors}, SLAM~\cite{Fox2012TactileRobot}, and Monte Carlo methods to estimate object pose. These are well-developed methods but, they may require multiple contacts with the object and may also suffer from slow run-time performance~\cite{Luo2017RoboticReview}.

Recently, deep learning approaches to 6D pose estimation have shown promising results. A differentiating factor between different approaches is the input of the network. Xiang et al.~\cite{Xiang2018PoseCNN:Scenes} proposed a novel network, PoseCNN, to estimate the 6D pose of an object. PoseCNN uses an RGB image to obtain an initial estimate of the 6D pose. Afterwards, depth sensor data is used to improve the initial estimate of the pose of the object using iterative closest point (ICP) algorithm.

Similar to PoseCNN, Tremblay et al.\cite{Tremblay2018DeepObjects} used a CNN trained on RGB images to obtain the $6$D pose estimate of an object. The authors used synthetic dataset to train their network. To bridge the sim-to-real gap, the authors developed a unique approach that uses a combination of domain randomized images and photo-realistic data to train the CNN. The authors present results that indicate a network trained on synthetic dataset can generalize to outperform the state-of-the-art on the YCB-Video dataset.

Wang et al.~\cite{Wang2019DenseFusion:Fusion} propose a method that uses both RGB images and depth data to estimate the 6D pose of an object. The authors propose a novel method to fuse the RGB and depth data. Instead of global level, the data are fused using pixel-wise fusion. The benefit of using pixel-wise fusion is that it allows a network to be trained on the visible part of the object, minimizing the effects of occlusion. 

Most of the literature on $6$D pose estimation, does not use tactile data. Recently, Bauza et al.~\cite{Bauza2020TactileRendering} presented a pose estimation approach that uses tactile sensors. Similar to Bimbo et al.~\cite{Bimbo2012ObjectInformation}, the authors use tactile data to estimate the local shape of the object at the contact point. The estimate of the local shape is then ranked against a set of simulated contacts, finding the probability distribution of a likely object pose that generated the contact surface. 

\begin{figure*}[t!]
    \centering
    \includegraphics[trim={0cm 0cm 0cm 0cm}, clip, width=0.85\textwidth]{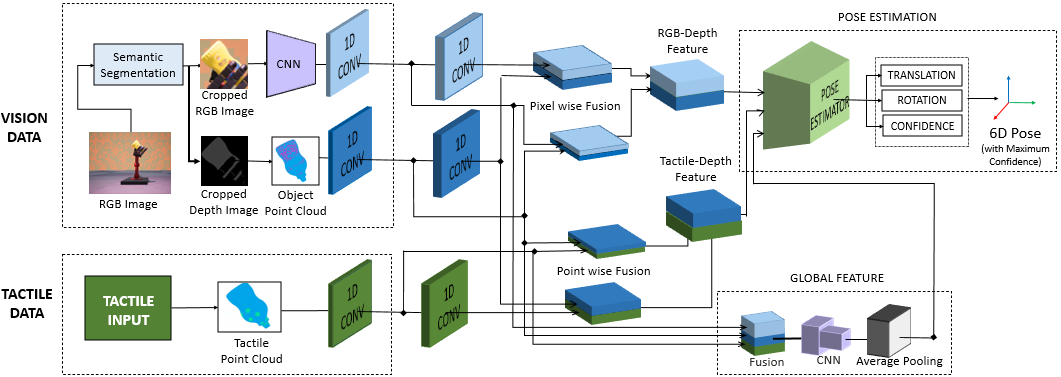}
    \caption{The proposed network architecture: given a color image, depth image and the object-surface point cloud at finger-object contact location, the network estimates the $6D$ pose of the object. See Fig.~\ref{fig:pose_estimator} for details of pose estimation network.\vspace{-0.4cm}}
    \label{fig:Network_layout}
\end{figure*}

Watkins et al.~\cite{Watkins-Valls2019} propose a framework to determine the geometric shape of unknown objects. In this framework depth and tactile data are used as the input of a CNN to estimate the object shape. The authors demonstrated adding tactile data improves the accuracy of the estimated shape.

The literature makes significant advances in shape and pose estimation. Watkins et al.~\cite{Watkins-Valls2019} show that using tactile data significantly improves shape estimation accuracy for occluded objects, however, they do not consider estimating the pose of the object. Most of state-of-the-art $6D$ pose estimation algorithms use either RGB data or both RGB and depth data, but none leverage tactile data, making them unsuitable for in-hand object manipulation where robot's grippers present heavy self occlusion. The work presented in this paper seeks to address the problem of $6D$ pose estimation for in-hand object manipulation.

\section{Network Architecture}
\label{sec:network-architecture}

In this section, we present a method to estimate the 6D pose of a known object that is in the grasp of a robot. The inputs to our model are color and depth images from a camera, and tactile contact points in the form of object-surface point cloud.
The output of our network is the 6D pose of the object in camera's coordinate frame.

\subsection{Visuo-Tactile Sensor Fusion}
Figure~\ref{fig:Network_layout} illustrates the overall architecture of our network. As described in Section~\ref{sec:introduction}, the color, depth and tactile input are fused in a way that maintains correspondence between features.  The sensor fusion is divided into two channels: the visual channel and the tactile channel. 

\subsubsection{The Visual Channel} 
the input to the visual channel consists of the color and the depth images from the camera. First, semantic object segmentation is performed on the color image to segment the object of interest. In semantic segmentation, each pixel is associated with an object class. In our experiments we adopted the method presented by Xian et al.~\cite{Xiang2018PoseCNN:Scenes}, which uses an encoder-decoder architecture. 

The depth image is masked using object's semantic segmentation and converted to a 3D point cloud in the camera's coordinate. That is, for each object-pixel visible in the color image, we associate the corresponding 3D location of the object in the camera's coordinate frame.

In the next step, the color image is cropped by the bounding box of the semantic segmentation. The cropped image is then fed into a convolution neural network (CNN) that processes the color image and maps each pixel in the cropped image to a color feature embedding. Since the color embedding is for the entire cropped image, the mask from the semantic segmentation is used to keep only $n$ random embeddings and their corresponding 3D points. This ensures that the subsequent networks only consider the points that are on the object. In our experiments $n$ was set to $1000$. 

The color embedding and the depth embedding, each goes through a two-stage CNN. The output of the color embedding CNN and the depth embedding CNN are fused using pixel-wise dense fusion. 

\subsubsection{The Tactile Channel} 
the input to the tactile channel consists of the point cloud from the camera's depth image and the object-surface point cloud from the tactile sensors. The tactile data from all fingers is presented as a single 3D point cloud to the network. This choice allows us to make the network invariant to the number of fingers a robot may have, i.e., making it gripper invariant. Similar to the vision channel, we only keep $n$ random points. The tactile data then goes through a two-stage CNN to produce tactile embeddings. The tactile embeddings and the depth embedding are fused using point-wise dense fusion.

\subsubsection{Global Feature}
in order to provide a global context to the network, we also generate a global feature by fusing the color, depth and tactile embeddings, followed by a CNN and an average pooling layer.

\subsection{The Pose Estimator}
\begin{figure}[t!]
    \centering
    \includegraphics[width=0.50\textwidth]{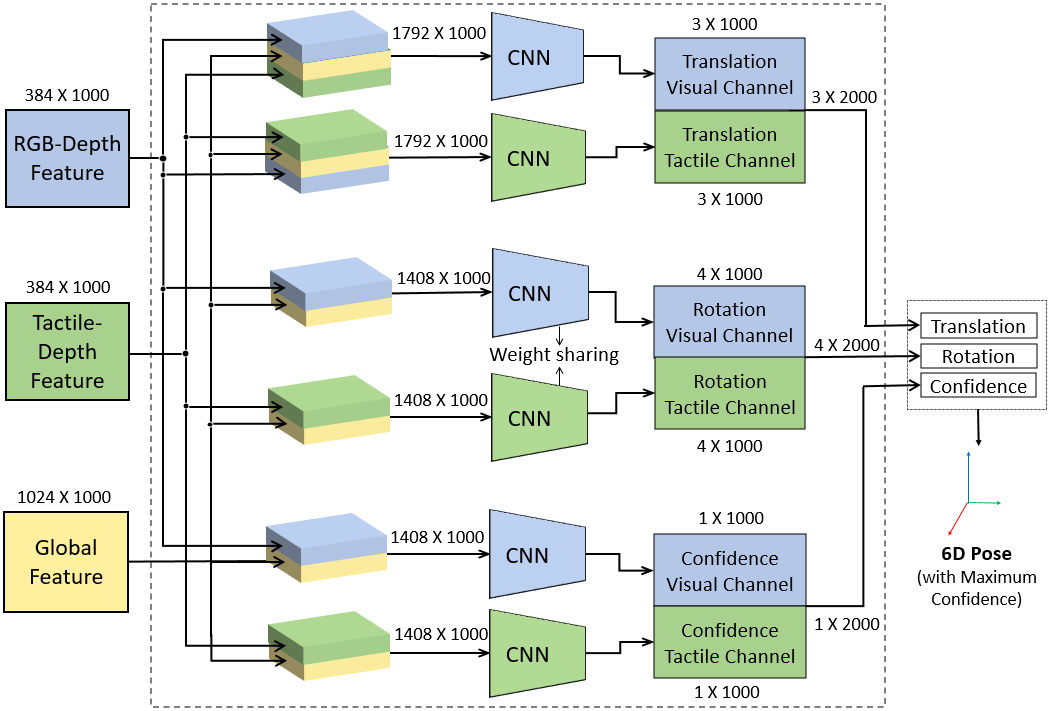}
    \caption{The network architecture for the pose estimator. The colors blue, green and yellow denote visual-channel features, tactile-channel features, and global feature, respectively.\vspace{-.7cm}}
    \label{fig:pose_estimator}
\end{figure}

The sensor data fused in the previous section are input to the pose estimator, namely, the RGB-depth features from the visual-channel, the tactile-depth features from the tactile-channel and the global feature. Figure~\ref{fig:pose_estimator} shows the pose estimator network architecture. The network estimates a translation vector, a rotation vector and a confidence level for each feature in the visual-channel and the tactile-channel, denoted by blue and green colors, respectively. To estimate the translation vector the RGB-depth features, tactile-depth features and the global feature are fed to a visual-channel CNN (blue) and a tactile-channel CNN (green). To estimate the rotation vector, the tactile-depth features combined with the global feature is fed to the tactile-channel CNN whereas the RGB-depth features combined with the global feature is fed to the visual-channel CNN in a Siamese network architecture, where the two networks share weights. The confidence levels are estimated by combining the tactile-depth features with the global feature through a tactile-channel CNN, whereas RGB-depth features combined with the global feature are fed to a visual-channel CNN. In our experiments, the visual and tactile channels, each, output 1000 pose estimates with the associated confidence levels. Since these points are randomly selected, depending on the location and the RGB color they vary in their accuracy of pose estimates. By learning a confidence level, the network chooses a pose estimate based on most informative feature point. Thereby, the 6D pose with the maximum confidence level is selected.

\subsection{The Loss Function}

We use the loss function in~(\ref{eq:loss_equation2}) to train our network. As mentioned in the previous section, for each point, $i$, in the input, $N$, the network estimates a translation vector, $\hat{T}_i$, a rotation vector, $\hat{R}_i$, and a confidence level, $C_i$. Given the model of the object with $M$ points, the point-wise loss, $L^p_i$, is calculated by transforming the model points by the estimated pose $\hat{p}=[\hat{R}_i,\hat{T}_i]$ and comparing it with the same model points transformed in ground truth pose $p=[R,T]$ as defined in~(\ref{eq:loss_equation1}). 

The point-wise confidence loss, $L^c_i$, is difference between the estimated confidence level, $\hat{C}_i$, and the expected confidence, $C_i^e$, and a regularization term $w log(C_i)$ as defined in~(\ref{eq:confidence_loss}). The regularization term penalizes the points with high confidence value. The expected confidence is the exponential function of pixel loss, $C_i^{e}= 0.5^{L^p_i}$. The network loss, $L$, is defined in~(\ref{eq:loss_equation2}) where the sum of the point-wise loss and the confidence loss are added and averaged over all points.\vspace{-.3cm}

\begin{align}
    \label{eq:loss_equation1}
    L^p_i &= \frac{1}{M} \sum_{j \epsilon M} ||(Rx_j + t) - (\hat{R}_ix_j + \hat{t}_i) ||\\
    \label{eq:confidence_loss}
    L^c_i&= (\hat{C}_i-C_i^{e} - w~log(C_i))\\
    \label{eq:loss_equation2}
    L &= \frac{1}{N} \sum_{i \epsilon N} (L^p_i + L^c_i)
\end{align}

\section{Dataset Generation}\label{tactile_dataset}
In this section we will explain how we generate the tactile data, followed by dataset collection setup and characteristics.

\subsection{Tactile Sensor Invariant Pose Estimation}\label{sec:tactile-invariant}

\begin{figure}[t!]
    \centering
    \begin{subfigure}[b]{0.35\textwidth}
      \centering
      \includegraphics[width=\textwidth]{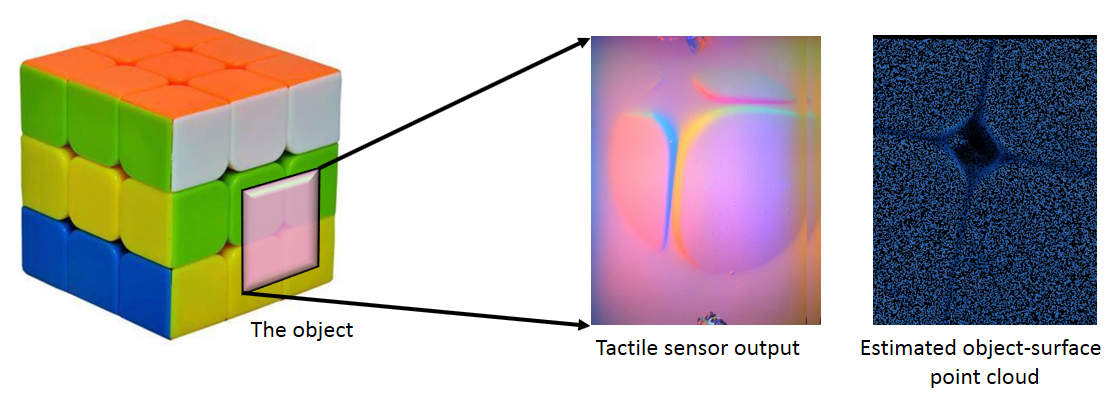}
      \caption{GelSight~\cite{Johnon2009GelSight} tactile sensors}
      \label{fig:gelsight-to-cloud}
    \end{subfigure}
    \centering
    \begin{subfigure}[b]{0.4\textwidth}
      \centering
      \includegraphics[width=\textwidth]{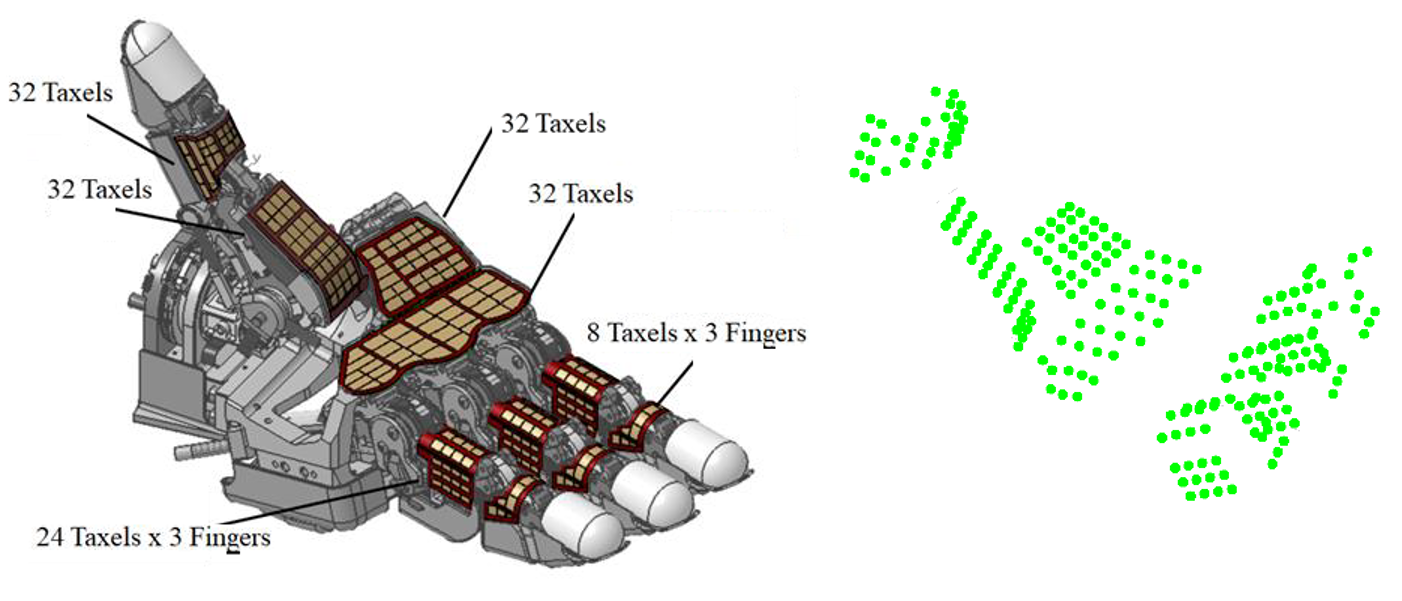}
      \caption{Multi-fingered hand with tactile sensors}
      \label{fig:rx-to-cloud}
    \end{subfigure}
    \caption{An illustration of tactile data to object surface point cloud generation: a) Optical-based tactile sensor and the corresponding point-cloud, b) Pressure-based tactile sensor and the corresponding tactile point-cloud.}
    \label{fig:tactile-point-cloud}
\end{figure}

As mentioned earlier, the spatial resolution of tactile sensors can vary greatly between different tactile sensors. 
To make the pose estimation algorithms invariant to tactile sensor types, we propose that the tactile data to be presented to the algorithm in the form of object-surface point cloud. That is, when the tactile sensors make contact with the object, we estimate the object surface in contact with the tactile sensor from the tactile sensor data. Forward kinematics can be used to get the 3D position of each taxel in contact with the object. Figure~\ref{fig:tactile-point-cloud} shows how this can be achieved for an optical-based tactile sensor and a pressure-based tactile sensor. In addition to sensors, this representation also makes the network invariant to the gripper type used.


\subsection{Data Collection Setup}

\begin{figure}[t!]
    \centering
    \begin{subfigure}[b]{0.24\textwidth}
      \centering
      \includegraphics[width=\textwidth]{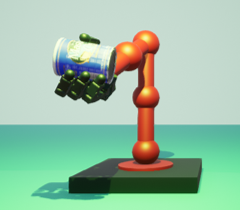}
      \caption{}
      \label{fig:unreal_dr-setup}
    \end{subfigure}
    \centering
    \begin{subfigure}[b]{0.20\textwidth}
      \centering
      \includegraphics[width=\textwidth]{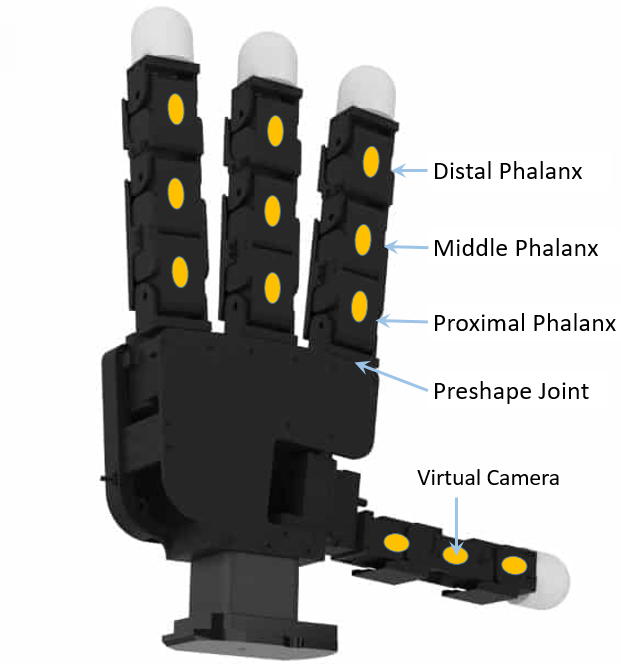}
      \caption{}
      \label{fig:takktile}
    \end{subfigure}
    \caption{a) Unreal Engine data collection setup with domain randomization b) The Allegro Hand with tactile sensors. The yellow ellipses show the location of virtual depth cameras used to generate tactile data in simulation.\vspace{-.04cm}}
    \label{fig:unreal-setup}
\end{figure}


\begin{figure*}[t!]
    \centering
    \begin{subfigure}[b]{0.3\textwidth}
      \centering
      \includegraphics[height=3.5cm]{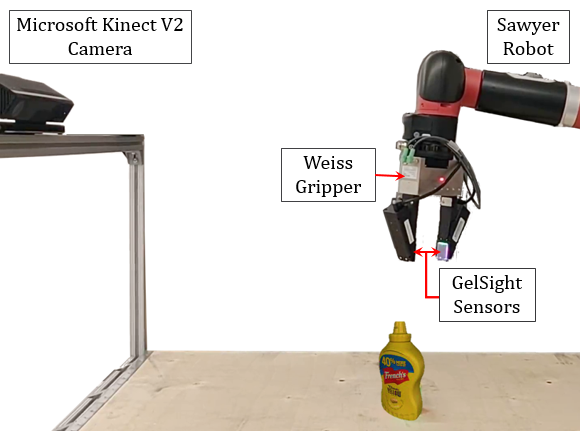}
      \caption{Robot Setup-I}
      \label{fig:hri-setup}
    \end{subfigure}
    \centering
    \begin{subfigure}[b]{0.3\textwidth}
      \centering
      \includegraphics[height=3.5cm]{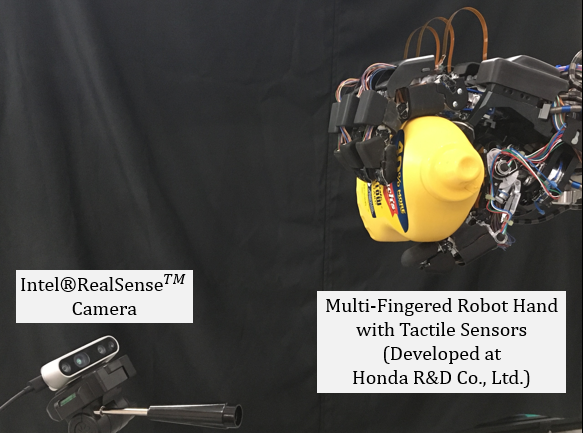}
      \caption{Robot Setup-II}
      \label{fig:rx-setup}
    \end{subfigure}
    \centering
    \begin{subfigure}[b]{0.38\textwidth}
      \centering
      \includegraphics[height=2.8cm]{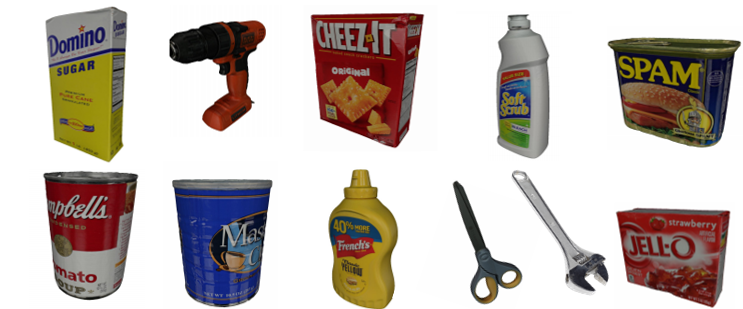}
      \caption{YCB objects}
      \label{fig:object_grid}
    \end{subfigure}
    \caption{Experimental setup for hardware deployment, and the YCB objects in our dataset.}
    \label{fig:hardware_deployment}
    \vspace{-0.3cm}
\end{figure*}


We adopted the approach proposed by Tremblay et al.~\cite{to2018ndds}, which uses the Unreal Engine 4 to produce photo-realistic simulated data. We extended the plugin developed by the authors to generate the tactile point clouds along with the color and depth data.

The simulation environment consists of a main camera, and a 6 DoF robotic arm with a 4-fingered gripper (Allegro Hand) equipped with tactile sensors. As shown in Fig.~\ref{fig:takktile}, we attach a virtual camera to each phalanx of the gripper. Each virtual camera's origin is on the surface of the respective phalanx, with the z-axis pointing out of the finger. These virtual cameras capture an object's surface point cloud when the gripper makes contact with the object. The main camera is placed in front of the robot, which generates the color and depth images. The dataset is generated by the procedure described in Algorithm~\ref{alg:data-collecction}.

\begin{algorithm}[t!] 
\caption{Data Collection Procedure}
\label{alg:data-collecction}
\begin{algorithmic}[1]
\ForEach {object}
\While {random trajectories $<$ 200}
\State place the object randomly in the robot's work-space
\State randomly set the object's orientation
\State randomly set the gripper preshape joint ( Fig.~\ref{fig:takktile})\;
\State grasp the object
\While {instances $<$ 100}
\State record RGB-D images,
\State record tactile object-surface point cloud
\State record object's pose in camera's coordinate
\State move the arm to a new position
\EndWhile
\State release the object
\EndWhile
\end{algorithmic}
\end{algorithm}





\subsection{Dataset Characteristics} 

We select 11  YCB objects~\cite{Calli2015TheResearch} shown in Fig.~\ref{fig:object_grid} based on the graspability and the quality of 3D models. The dataset comprises of 20000 samples per object, each consisting of depth images from the 12 virtual tactile cameras placed at every phalanx of the robotic gripper (Fig.~\ref{fig:takktile}) and RGB-D images from the main camera. The height to depth resolution of the camera placed on the fingers is 32x32 and is proportional to the shape of the tactile sensors. The main camera captures a 640x480 image of the entire scene. 

In order to provide sufficient variation in our dataset to improve network generalization, the collected dataset includes domain randomized samples. Twenty percent of the dataset consists of samples where no domain randomization is applied.  The remaining 80\% consists of domain randomized samples. We use the NVIDIA's Deep Learning Dataset Synthesizer's domain randomization plugin for Unreal Engine 4 to randomized the background, the floor surface, add a spotlight with randomized position and a color as shown in Fig.~\ref{fig:unreal_dr-setup}. 

\vspace{-0.25cm}
\section{Experiments}
The network is trained on the dataset described in Section~\ref{tactile_dataset}. 
\vspace{-0.2cm}
\subsection{Network Training Setup} 
The dataset is split into a 4:1 ratio with the training set consisting of 16000 samples and the test set consisting of 4000 samples for each object. The network is trained for 1500 epochs with a batch size of 16. The learning rate, \emph{lr}, for the network is set to $0.0002$. The hyper-parameter, \emph{w}, in~(\ref{eq:confidence_loss}), is set to $0.015$. The network was trained using Adam optimizer in PyTorch.
 

\subsection{Hardware Deployment}\label{subsec:hardware}
\label{sec:hardware-deployment}

We deployed our network on two different setups with different grippers and tactile sensors. The first setup (Setup-I) consists of a Weiss gripper with two Gelsight tactile sensors attached to a Sawyer robot. A Kinect2 RGB-D camera is mounted  $4\,m$ from the base of the robot, at $45^{\circ}$ facing the work-space (see Fig.~\ref{fig:hri-setup}). The work-space is instrumented with OptiTrack motion capture system to record the ground-truth 6D pose of the object of interest.

The second setup (Setup-II), as illustrated in Fig.~\ref{fig:rx-setup} consists of an Intel\textregistered RealSense\textsuperscript{TM} camera, and a multi-fingered robot hand developed at Honda R\&D Co. Ltd. shown in Fig.~\ref{fig:rx-to-cloud}. It has 4 fingers, 16 joints (11 actuated). It is also equipped with a tactile sensor suite of 224 taxels.

The network was deployed on a workstation with NVIDIA Quadro RTX 6000. The inference time for our network alone is 10.1 ms and, the entire pipeline, including the semantic segmentation network, and ROS overhead, takes 109.5 ms.

\section{Results}

\begin{figure}[t!]
    \centering
    \begin{subfigure}[b]{0.50\textwidth}
       \centering
       \includegraphics[width=\textwidth]{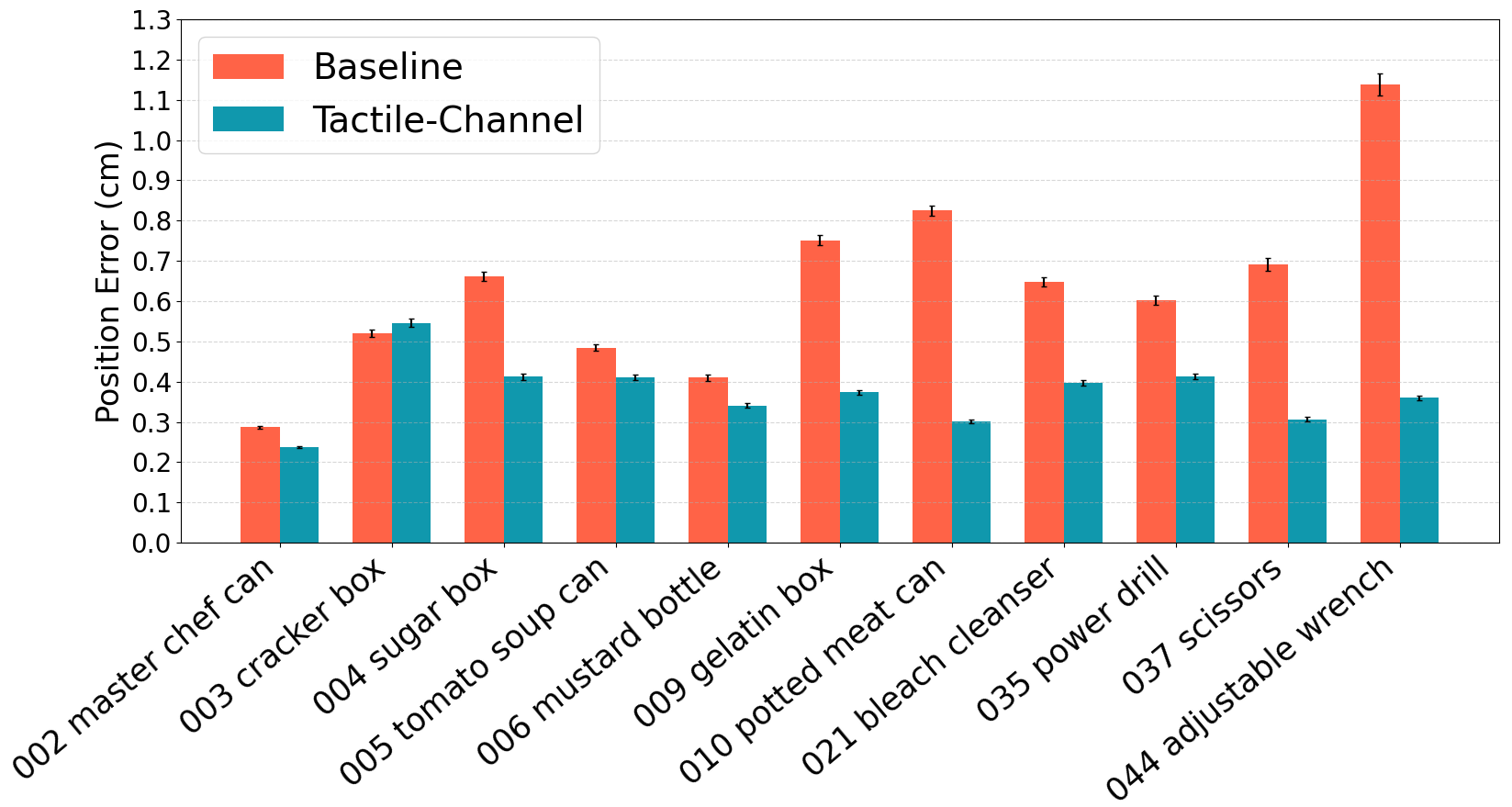}
       \caption{Position error}
       \label{fig:position_error}
    \end{subfigure}
    \centering
    \begin{subfigure}[b]{0.50\textwidth}
       \centering
       \includegraphics[width=\textwidth]{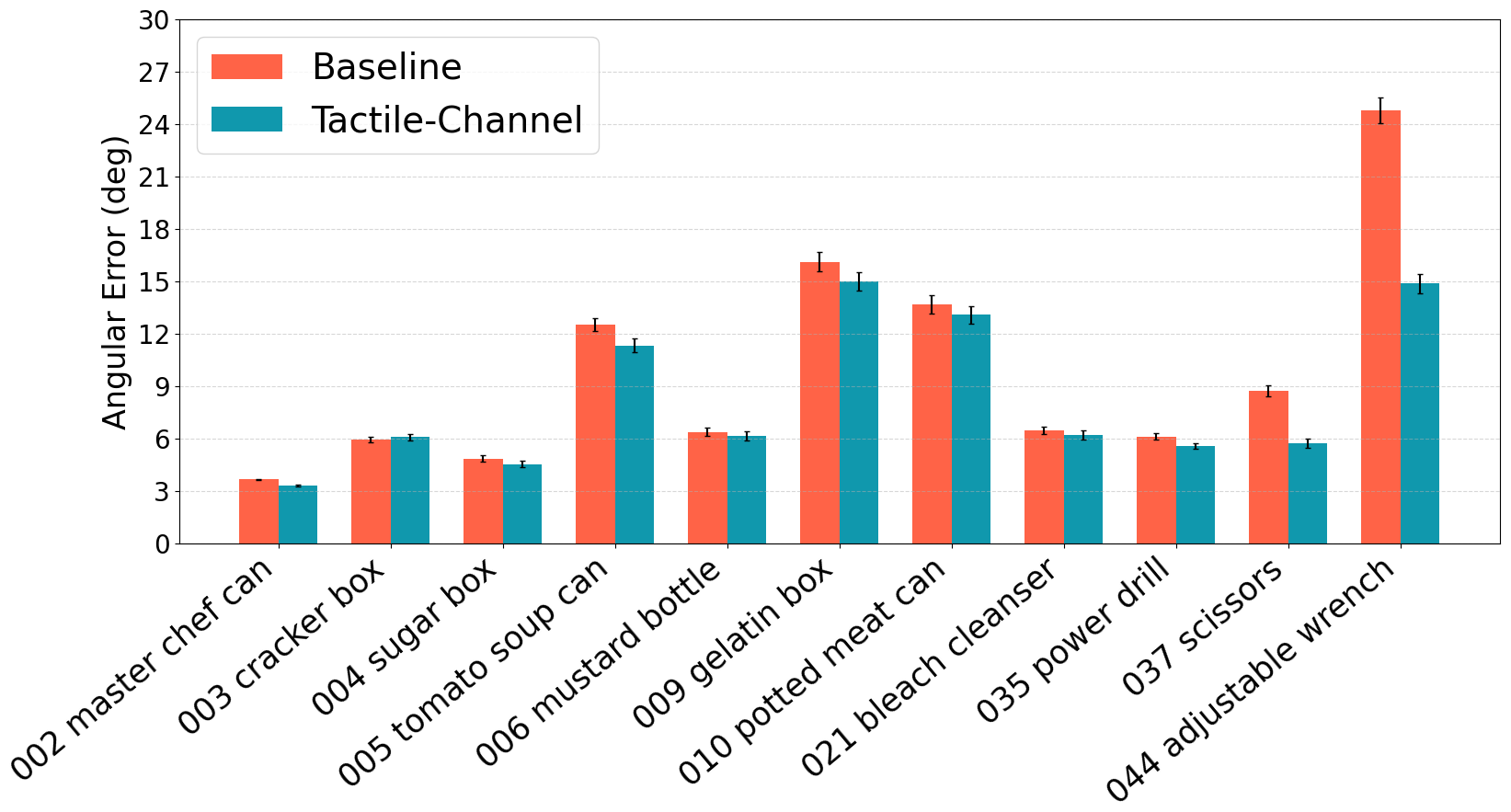}
       \caption{Angular error}
       \label{fig:angular_error}
    \end{subfigure}
    \caption{Position error and angular error for each object in our dataset. \vspace{-0.5cm}}
    \label{fig:position_angular_error}
\end{figure}




To test our hypothesis that using both tactile and vision data improves the 6D pose of an object, we present the results of experiments in which we compare our method with a baseline method that uses vision data alone to estimate the 6D pose of the object. The baseline method is constructed by modifying our network in which we remove the tactile channel, keeping only the visual channel. We then train this vision only network with color and depth images from the same dataset. In the following section, we present quantitative analysis of the performance of our method on the synthetic dataset, this is followed by qualitative analysis of the network deployed on a physical robot. All reported errors, unless otherwise stated, are standard errors.

%

\subsection{Quantitative Evaluation}
\label{sec:quantitative_evaluation}

To evaluate the performance of the proposed method we use two metrics: position error and angular error. The position error is the Euclidean distance between the estimated position, $\hat{T}$, and the ground truth position $T$. We use the angle between two unit quaternions to represent the angular error~\cite{Huynh2009MetricsAnalysis}, shown in (\ref{orientation_error}).\vspace{-0.3cm}

\begin{equation}
    \theta= \cos^{-1}(2\langle \hat{q} ,q \rangle^2-1)
    \label{orientation_error}
\end{equation}

where $\theta$ is the angular error, $\hat{q}$ is estimated quaternion from the network, $q$ is the ground truth quaternion, and $\langle \hat{q} ,q \rangle$ is the inner product of the two quaternions.

\subsubsection{Performance on Position and Angular Metrics}
Figure~\ref{fig:position_angular_error} shows the position error and the angular error for each object in our dataset. The results show for all objects, except cracker box, our network outperforms the baseline method with statistical significance. In some objects, for example, the adjustable wrench, the position error for our approach is $0.36\,cm$, compared to $1.14\,cm$ for the baseline. Other objects that exhibit significant improvement compared to the baseline include scissors, potted meat can and gelatin box. When analysing the results on angular data, we notice significant improvement for  adjustable wrench and for the scissors. We also notice, for angular estimates, the spread between the baseline and our methods is not as pronounced as the position estimates. These results suggest that the network relies on the color features obtained from the camera image to infer the orientation of the object. This might lead to visually dense objects like cracker box getting high confidence resulting in orientation estimates akin to the baseline and relatively less accurate position estimates.






\begin{figure}[t!]
    \centering
    \begin{subfigure}[b]{0.368\textwidth}
       \centering
       \includegraphics[width=\textwidth]{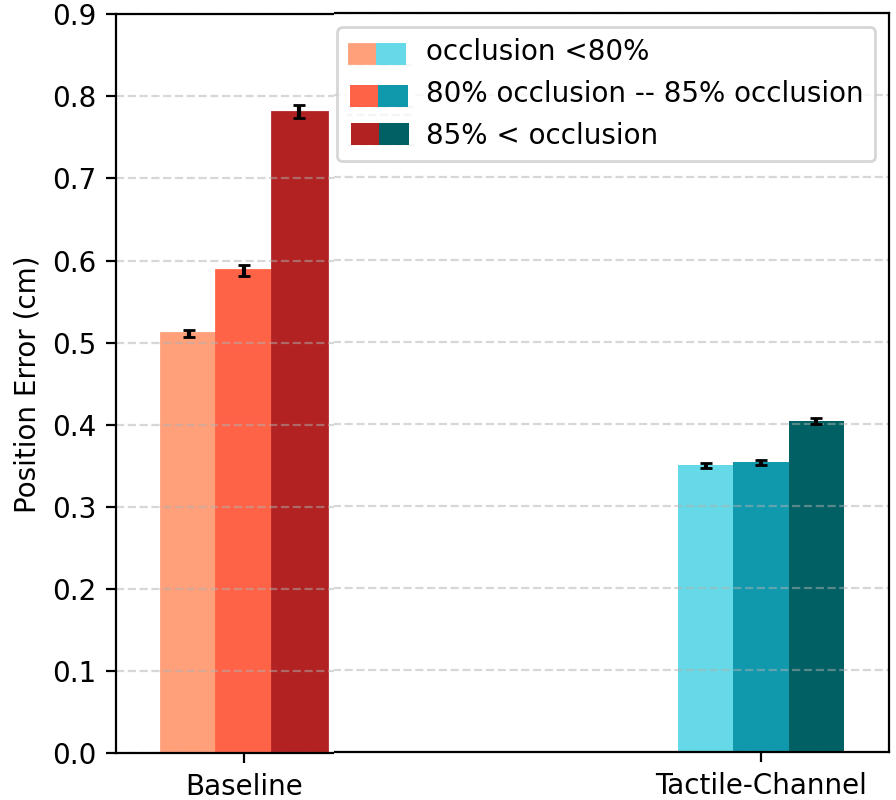}
       \caption{Position error}
       \label{fig:occ_position_error}
    \end{subfigure}
    \centering
    \begin{subfigure}[b]{0.368\textwidth}
       \centering
       \includegraphics[width=\textwidth]{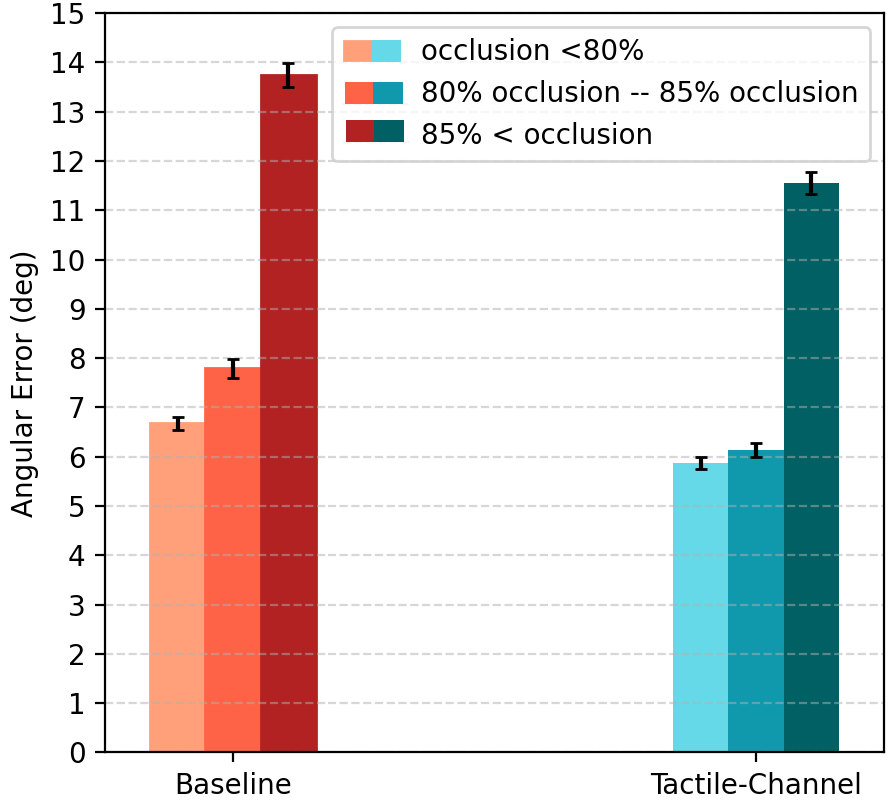}
       \caption{Angular error}
       \label{fig:occ_angular_error}
    \end{subfigure}
    \caption{Position and angular errors of the baseline and Tactile-Channel network at different occlusion levels. The darker shades represent higher level of occlusion.\vspace{-0.55cm} } 
    
    
    \label{fig:position_angular_occlusion}
\end{figure}

\subsubsection{Effect of Occlusion Level on Performance}
To analyse the effect of occlusion level on the performance of the network, we run an analysis in which we divided the performance of the network into three occlusion levels: below 80\%, between 80\% and 85\%, and above 85\%. Given 3D model of the object as a point cloud, and the point cloud obtained from the depth image of the main camera, occlusion level is calculated as the percentage of points in the model of the object that are visible to the main camera. First, we apply the ground truth transformation to the model, then, each point in the model point cloud is labelled visible if there exists a point within a threshold, $\epsilon$, in the main camera point cloud using the K-nearest Neighbor algorithm. The occlusion level is the percentage of points labelled as not visible in the model.

Figure~\ref{fig:position_angular_occlusion} shows the results of this analysis. We see significant drop in the performance of the baseline network as we increase the level of occlusion. Under heavy occlusion, the average position error for our network ($0.4\pm0.003$ cm) is almost half of the baseline ($0.78\pm0.008$ cm). We also see a reduction of $2.3$ degrees in the angular error where we get $11.5\pm0.22$ degrees, and $13.8\pm0.24$ degrees, for our network and the baseline, respectively. We notice compared to position error, in which our network maintains performance, the angular error under heavy occlusion, albeit outperforming the baseline, suffers a large degradation. This suggest that object texture is important for the inference of the orientation of an object. This is expected, for example, inferring whether the back or the front of an object like mustard bottle is facing the camera is difficult without any visual cues. In fact, looking at per object angular error (Fig.~\ref{fig:angular_error}), we notice the objects that demonstrate large angular performance gain when tactile data is included are those with less texture, for example, the adjustable wrench has uniform texture. Telling apart one side of the object to the other depends heavily on the shape of the object captured by the point cloud data.

\begin{figure}[t!]
    \centering
    \begin{subfigure}[b]{0.23\textwidth}
       \centering
       \includegraphics[width=\textwidth]{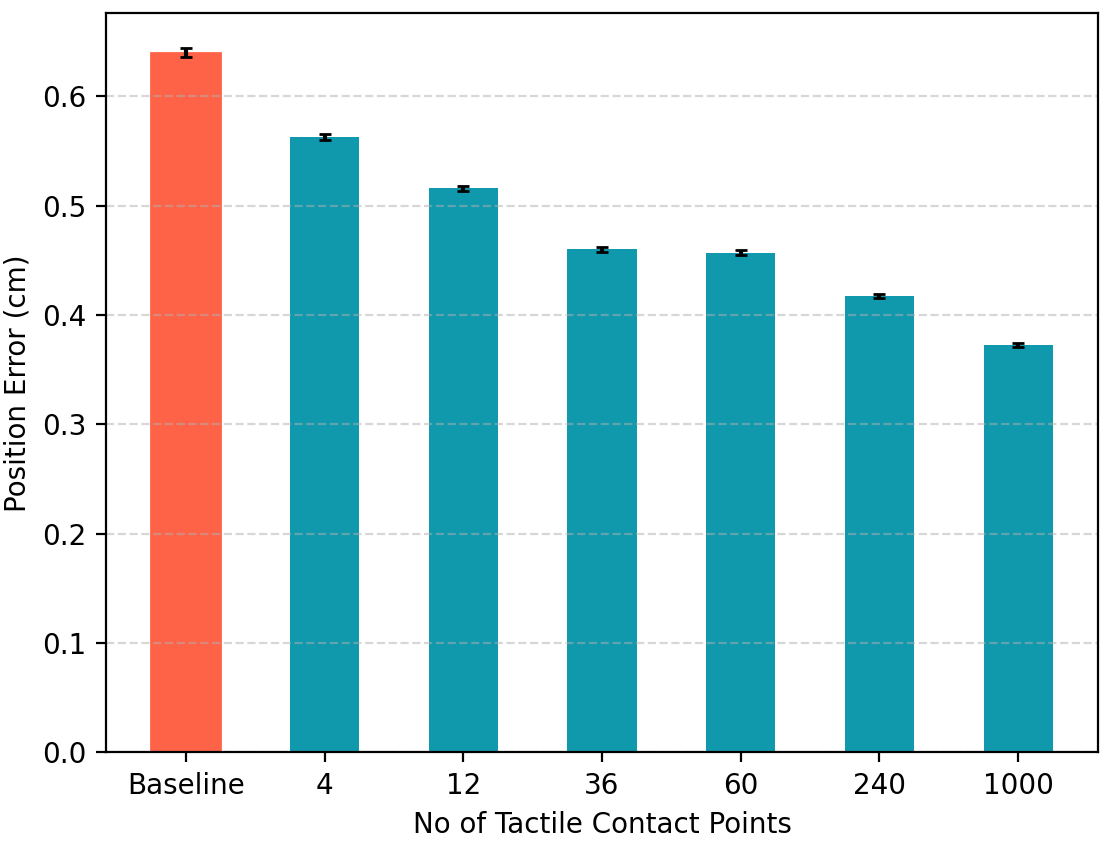}
       \caption{Position error}
       \label{fig:variable_tactie_position_error}
    \end{subfigure}
    \centering
    \begin{subfigure}[b]{0.23\textwidth}
       \centering
       \includegraphics[width=\textwidth]{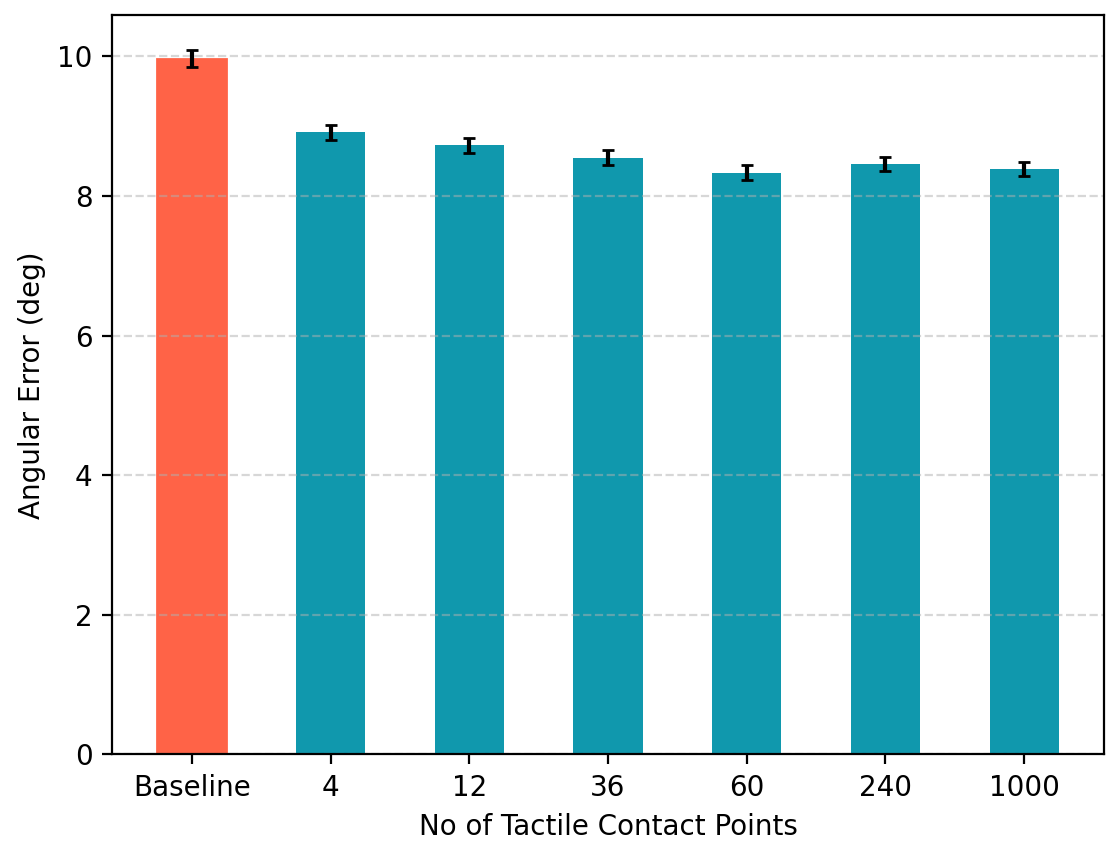}
       \caption{Angular error}
       \label{fig:variable_tactile_angular_error}
    \end{subfigure}
    \caption{Effect of varying tactile contact points on the network performance, where \baselinecolor and \tactilechannelcolor bars represent the baseline and the Tactile-Channel networks, respectively.\vspace{-0.4cm}}
    
    
    \label{fig:variable_tactile}
\end{figure}

\subsubsection{Effect of Number of Tactile Points on The Performance}
We run further analysis to study the effect of the number of the tactile contact points on the performance. We modified the proposed network to accept 4, 12, 36, 60, 240 tactile contact points. Figure~\ref{fig:variable_tactile} shows the results of this experiment for all objects in our dataset. We also plot the performance of the baseline network for reference. As expected a decrease in the number of tactile contact points increases the position and the angular error. The results suggest inclusion of tactile data is an improvement over a vision only baseline even if the number of tactile contact points is reduced.





\subsection{Qualitative Evaluation}
\label{sce:qualitative_evaluation}

In this section we present qualitative analysis of deploying our network and the baseline network simultaneously in real-time on the hardware setup described in section~\ref{subsec:hardware}. Note, that the networks are trained on the synthetic dataset only. During deployment, both our network and the baseline get the same input. The accompanying video is a visualization of the results presented in this section. The average position and angular error for the data presented in the video for the Robot Setup-I (Fig. \ref{fig:hri-setup}) are $36.88\pm1.27$ degrees and $2.69\pm0.13$ cm respectively, for our network, compared to $64.57\pm1.40$ degrees and $4.76\pm0.32$ cm for the baseline.


\subsubsection{Performance on Position and Angular Metrics}
\begin{figure*}[t!]
    \centering
    \begin{subfigure}[b]{0.65\textwidth}
    \centering
    \includegraphics[height=2.7cm]{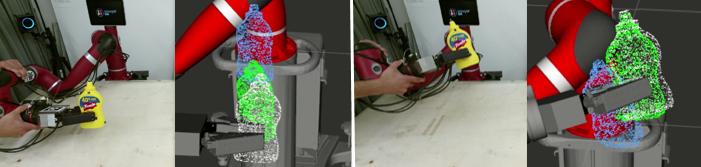}
    \caption{Setup-I}
    \end{subfigure}
    \centering
    \begin{subfigure}[b]{0.32\textwidth}
    \centering
    \includegraphics[height=2.7cm]{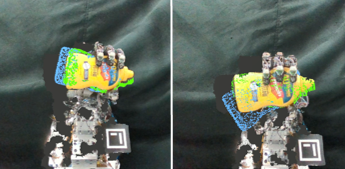}
    \caption{Setup-II}
    \end{subfigure}
    \caption{Deployment on setups described in Section~\ref{sec:hardware-deployment}: the green and the blue point clouds represent the Tactile-Channel and the baseline, respectively. Setup-I, left image is the view from the main camera, in the right image the white point cloud represents the ground truth pose from the motion capture system. Note, Setup-II does not have a motion capture system.\vspace{-0.3cm}}
    \label{fig:video_screenshots}
\end{figure*}

During deployment our network demonstrated stable frame-by-frame output compared to the baseline network.  Figure~\ref{fig:video_screenshots} shows examples of instances in which the baseline (blue) network's estimate deviates significantly from the ground-truth (white) pose, where our network (green) is able maintain reasonable estimate. This suggest that the network is able to use the tactile information, especially, from the fingers at the back of the object to get a better estimate of the object's 6D pose.


\subsubsection{Effect of Occlusion Level on the Performance}

\begin{figure*}[t!]
    \centering
    \includegraphics[width=0.7\textwidth]{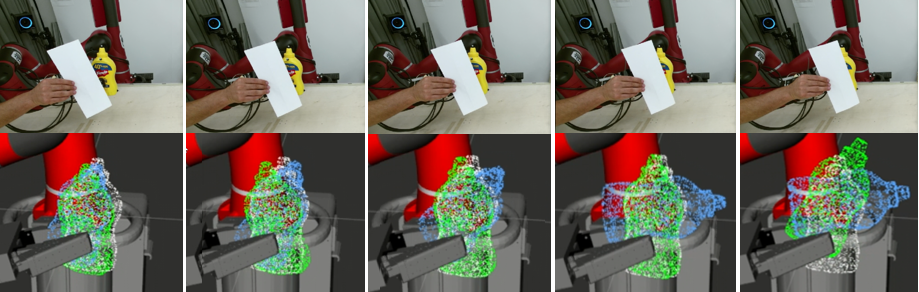}
    \caption{Pose estimation under different occlusion levels on the robot Setup-I described in Section~\ref{sec:hardware-deployment}, where white, green and blue point clouds represent the ground truth, our network and the baseline, respectively.\vspace{-0.3cm}}
    \label{fig:robot_occlusion}
\end{figure*}

We also run an experiment on the real hardware setup in which we gradually increased the occlusion level by blocking main camera's view using a piece of paper. Figure~\ref{fig:robot_occlusion} shows results of this experiment, demonstrating that our network (green) can handle higher levels of occlusion compared to the baseline network (blue). The ground-truth pose is shown in white. The results suggest that under heavy occlusion a network that uses both tactile and vision data is an improvement over a network that uses vision only.

\subsection{Ablation Study}

To evaluate the effect of our design choices on the performance of the network, we conducted three ablation studies to quantify the gain from: 1) the Siamese network, 2) the global feature and 3) the visual features, i.e., a Tactile-Only network by removing the visual channel and training on tactile data only. Table~\ref{table:ablation} shows the average network performance for all 11 objects. The results suggest that use of Siamese network results in a gain of $0.006\,cm$ in position and $0.2$ degrees in angular accuracy. The global feature contributes a $0.57\,cm$ gain in position and a $3.31$ degree  gain in angular accuracy. The visual data contributes a $1.55\,cm$ gain in the position accuracy, and a $51.19$ degree gain in the angular accuracy, which is consistent with our observation that inference of the orientation of an object is dependent on the color features from the RGB image.

\begin{table}[t!]
\caption{Ablation Study Results}
\centering
\begin{tabular}{ |c|c|c|}
 \hline
 Model & Position Error (cm) & Angular Error (deg) \\
 \hline
Tactile-Channel(Siamese) & 0.299~$\pm$~0.002 & \,\,\,8.074~$\pm$~0.105 \\
Non-Siamese & 0.305~$\pm$~0.002 & \,\,\,8.269~$\pm$~0.102 \\
No Global Features & 0.876~$\pm$~0.004 & 11.385~$\pm$~0.123 \\
Tactile-Only & 1.849~$\pm$~0.009 & 59.268~$\pm$~0.282\\
 \hline
\end{tabular}
\label{table:ablation}
\end{table}

\subsection{Comparison with State-of-the-Art}

\begin{table}
\caption{Comparison with State-of-the-Art}
\centering
\begin{tabular}{ |c|c|c|}
 \hline
 Model & Position Error (cm) & Angular Error (deg) \\
 \hline
PoseCNN & 6.146~$\pm$~0.023 & 10.896~$\pm$~0.082 \\
PoseCNN+ICP & 6.158~$\pm$~0.023 & 10.897~$\pm$~0.083 \\
Tactile-Channel & 0.299~$\pm$~0.002 & \,\,\,8.074~$\pm$~0.105 \\
 \hline
\end{tabular}
\label{table:posecnn}
\end{table}

To assess the difficulty level of our dataset we trained PoseCNN~\cite{Xiang2018PoseCNN:Scenes}, a vision based state-of-the-art 6D pose estimation algorithm on our dataset. Table~\ref{table:posecnn} shows the results of this analysis, we notice a significant performance degradation in the position estimate, which is expected for a method that uses only RGB images with heavy occlusion from a robot's grippers. We also notice a 4.78 degree increase in the angular error compared to our method. These results suggest our dataset presents a challenging pose estimation problem.


\section{Conclusion and Future work}

We proposed a novel approach to solve the problem of estimating $6D$ pose of an in-hand object. A challenge in in-hand pose estimation is how to compensate geometric information around the largely occluded parts of the object. In the proposed approach, we take advantage of the tactile sensing data on the robot fingers, which occlude the object.  We proposed a sensor agnostic representation in which tactile data is represented by finger-object contact-surface point cloud. We introduced a visuotactile network architecture in which data from vision and tactile sensors are fused in a meaningful way. To efficiently train our model, we proposed a synthetic data collection procedure to generate a large scale visuotactile dataset suitable for in-hand $6D$ pose estimation. We presented the results of experiments in which we compared our approach with a vision only baseline. Our results on the synthetic data suggest that our network outperforms the vision only baseline with statistical significance, especially under heavy occlusion. We also presented results of a study in which we varied the number tactile inputs between 4 and 240. Our results suggest, while there is a degradation in the accuracy of the estimates, our approach outperforms the baseline even when the number of tactile input is reduced to 4. Furthermore, we presented qualitative results in which we deployed our network on real physical robots with different hardware setups. Our qualitative results show that a network trained only on the synthetic data can successfully transfer to real world settings. 

One of the limitations of our approach is that orientation inference has a dependency on the object's color features from the vision sensor. While our method maintains performance when there is moderate to heavy occlusion, under very heavy occlusion, the performance degrades to that of the baseline. To alleviate this limitation a number of approaches can be pursued as future work. One approach is to use methods such as long short-term memory networks to use temporal coherence to filter out erroneous estimates. Another interesting direction is to utilize the geometry of the gripper and the position of the fingers to help the network filter out physically implausible pose estimates. Most of the objects in our dataset are symmetrical at least on one axis. In future we would like to extend our dataset to include 3D printed asymmetrical objects to evaluate the efficacy of our method on such objects. We also noticed that our method is robust to environmental occlusions. In future, we would like to conduct a study to test our method against different levels of environmental occlusions.

\vspace{-0.3cm}

\ifCLASSOPTIONcaptionsoff
  \newpage
\fi



\bibliographystyle{IEEEtran}
\bibliography{./bibliography/IEEEabrv,./bibliography/references.bib}
\end{document}